\def\year{2022}\relax
\documentclass[letterpaper]{article} 
\usepackage{aaai22}  
\usepackage{times}  
\usepackage{helvet}  
\usepackage{courier}  
\usepackage[hyphens]{url}  
\usepackage{graphicx} 
\usepackage{natbib}  
\usepackage{caption} 
\DeclareCaptionStyle{ruled}{labelfont=normalfont,labelsep=colon,strut=off} 
\nocopyright
\title{Generative Structured Normalizing Flow Gaussian Processes Applied to Spectroscopic Data}
\author{
    Natalie Klein, 
    Nishant Panda, 
    Patrick Gasda, 
    Diane Oyen
}
\affiliations{
    Los Alamos National Laboratory\\

    Los Alamos, New Mexico 87545\\
    neklein@lanl.gov, nishpan@lanl.gov, gasda@lanl.gov, doyen@lanl.gov
}

\usepackage{amsmath}
\usepackage{amssymb}

\begin{document}

\maketitle

\begin{abstract}
In this work, we propose a novel generative model for mapping inputs to structured, high-dimensional outputs using structured conditional normalizing flows and Gaussian process regression.
The model is motivated by the need to characterize uncertainty in the input/output relationship when making inferences on new data.
In particular, in the physical sciences, limited training data may not adequately characterize future observed data; it is critical that models adequately indicate uncertainty, particularly when they may be asked to extrapolate.
In our proposed model, structured conditional normalizing flows provide parsimonious latent representations that relate to the inputs through a Gaussian process, providing exact likelihood calculations and uncertainty that naturally increases away from the training data inputs.
We demonstrate the methodology on laser-induced breakdown spectroscopy data from the ChemCam instrument onboard the Mars rover Curiosity.
ChemCam was designed to recover the chemical composition of rock and soil samples by measuring the spectral properties of plasma atomic emissions induced by a laser pulse.
We show that our model can generate realistic spectra conditional on a given chemical composition and that we can use the model to perform uncertainty quantification of chemical compositions for new observed spectra.
Based on our results, we anticipate that our proposed modeling approach may be useful in other scientific domains with high-dimensional, complex structure where it is important to quantify predictive uncertainty.
\end{abstract}

\section{Introduction}
Physical systems can often be characterized by structured, high-dimensional measurements with variation depending on relatively simple system inputs.
Supervised learning models often successfully learn to predict system inputs from observed data (e.g., in spectroscopic data \cite{sun2019machine, ho2019rapid, castorena2021deep}).
However, such data-driven models are often trained using simulated or lab experiment data that is intended to cover the space of possibilities that will be encountered in the wild -- but what happens in the likely case that new observed data deviates from the training distribution?
Unfortunately, predictive algorithms often fail silently when asked to extrapolate to unobserved regions of data space.
In this paper, we develop a novel generative model approach that addresses this issue by capturing uncertainty in the relationship between system inputs and outputs. 
We demonstrate the utility of our approach on a real spectroscopic data set.

Our proposed approach posits a Gaussian process (GP) on the input space that predicts high-dimensional, complex structure in the output space through a structured conditional normalizing flow.
That is, the GP model represents variation in low-dimensional representations of the outputs as uncertain functions of the inputs.
We term the model a structured normalizing flow Gaussian process (SNFGP). 
Importantly, SNFGP models have the property that predictive uncertainty increases when a query input is far from the training inputs, helping to indicate when the model is being asked to extrapolate.
We show on a real spectroscopic data set that the SNFGP model performs well as a conditional generative model and that it provides a principled way to estimate model inputs with uncertainty, even in the extrapolation regime.

Specifically, we demonstrate our approach on experimental ChemCam calibration data~\cite{maurice2012chemcam}.
ChemCam measures laser-induced breakdown spectroscopy (LIBS), a type of atomic emission spectroscopy in which the laser makes a small plasma on the target surface, causing the atoms to emit light that is collected by high-resolution spectrometers.
The primary driver in variation across spectra is the chemical composition of the target, so previous works sought to predict the composition from the spectrum using methods like linear regression, dimension reduction, and deep neural networks  \cite{wiens2013pre,forni2013independent, clegg2017recalibration, anderson2017improved, castorena2021deep}.
However, these supervised learning approaches are, out of necessity, trained on calibration data collected from an instrument in the laboratory on Earth. 
When applied to data collected on Mars, there is a risk of extrapolation due to unanticipated compositions, suggesting that estimates of uncertainty are crucial for interpretation of results.
Our results indicate that in this data set, the SNFGP model achieves good fit, generates accurate distributions over ChemCam spectra conditional on an input, and provides an avenue for estimating system inputs (i.e., chemical compositions) with meaningful uncertainties. 
We anticipate that the SNFGP model could be useful in many scientific areas with high-dimensional, structured features.

\section{Related Work}
Normalizing flows (NFs) have found success as generative models compared to alternatives like generative adversarial networks or variational autoencoders (VAEs) \cite{rezende2015variational, papamakarios2019normalizing}.
Traditional normalizing flows describe the data distribution, but do not condition on inputs.
Recently, \cite{winkler2019learning} presented conditional NFs with a diagonal Gaussian distribution conditional on inputs $\mathbf{x}$ which, unlike SNFGP, does not explicitly model correlation across observations (key for estimation of extrapolation uncertainty).
A separate recent line of work in deep generative models uses GPs along with VAEs, either by using the GP as a prior in the variational objective or using a structured approximate posterior incorporating the GP \cite{casale2018gaussian,pearce2020gaussian}.
However, because VAEs are not comprised of invertible transformations, the exact likelihood is not available, complicating usage of the model for inverse uncertainty quantification.
Most similar to our work is \cite{maronas2021transforming} which discussed combining normalizing flows with Gaussian processes, but focused on applying normalizing flows to the Gaussian process prior rather than to the output of the Gaussian process. 
In that sense, our work is more similar to warped GPs \cite{snelson2004warped} but with dimension reduction and more flexible mappings.
By including dimension reduction, our work rests on earlier work in the statistics literature that used GPs in a principal components latent space for uncertainty quantification \cite{higdon2008computer}.

\section{Background}
\subsection{ChemCam Spectroscopy Data}
Since landing at Gale Crater in 2012, the ChemCam LIBS instrument onboard the Mars rover Curiosity has obtained spectral measurements of thousands of rock and soil analysis targets~\cite{maurice2016chemcam}. 
In this work, we use calibration spectra: LIBS spectra with corresponding known compositions collected in the laboratory on Earth (using an instrument very similar to ChemCam inside a chamber mimicking Mars' atmospheric pressure and composition).
The ChemCam spectral measurements come from three spectrometers (UV, VIO, and VNIR ranges). 
We use data coming from wavelength ranges of [246, 338], [382, 473], and [492, 849] nanometers, respectively, resulting in a total of 5,205 measured wavelengths.
To demonstrate our method, we focus on the oxide weight percent of SiO$_2$ as a simplified (though incomplete) description of the composition of each material.
For each spectrum, we normalize the intensity values within each spectrometer to the sum of all spectral values within the spectrometer range.
In total, we have a set of 2,442 calibration spectra corresponding to 466 unique materials.

\subsection{Gaussian Process Regression}
Gaussian process regression models the distribution of random variables $Z$ conditional on $X$. 
For simplicity, we assume $D$-dimensional observed input data vectors $\{\mathbf{x}^{(i)}\}_{i=1}^N$ and corresponding univariate output data $\mathbf{z} = \{z^{(i)}\}_{i=1}^N$.
The data are modeled as zero-mean Gaussian jointly with unobserved function values $z^*$ associated with a new input $\mathbf{x}^*$:
\begin{align}
    \begin{bmatrix}
    \mathbf{z} \\
    z^*
    \end{bmatrix} \sim 
    \mathcal{N} \left( 0, 
    \begin{bmatrix}
    \Sigma_\mathbf{z} + \sigma^2_\epsilon \mathbf{I} & \Sigma_{\mathbf{z}, z^*} \\
    \Sigma_{z^*, \mathbf{z}} & \Sigma_{z^*}
    \end{bmatrix}
    \right). \label{eq:gpjoint}
\end{align}
In Eq~\ref{eq:gpjoint}, $\sigma^2_\epsilon$ represents the variance of additive Gaussian white noise on the observed data.
The covariance matrices are parameterized through covariance functions dependent on hyperparameters $\boldsymbol{\gamma}$; in this work, we use a covariance function that specifies entries of the covariance matrix as
\begin{align}
    \left[\Sigma_{\mathbf{z}}\right]_{i, j} = \sigma^2 \exp \left( - \sum_{d=1}^D \tfrac{\left(x^{(i)}_d - x^{(j)}_d\right)^2}{2 \ell_d^2} \right).
\end{align}

The log marginal likelihood of $\mathbf{z}$ (based on Eq~\ref{eq:gpjoint}) can be used to tune the hyperparameters $\boldsymbol{\gamma} = [\sigma^2, \ell_1, ..., \ell_D]$.
Using classic results on multivariate Gaussian distributions, the distribution of $z^*$ conditional on the observed data provides a predictive Gaussian distribution over $z^*$ given inputs $\mathbf{x}^*$.

\subsection{Normalizing Flows}
Normalizing flow models seek to model the data probability density function $p_Y(\mathbf{y})$ through a function $f_\phi$ applied to random variates from a simpler density function $p_Z(\mathbf{z})$. 
Using the change of variables formula:
\begin{align}
    p_Y(\mathbf{y}) = p_Z(f_\phi(\mathbf{y})) \left| \frac{\partial f_\phi(\mathbf{y})}{\partial \mathbf{y}} \right|.
\end{align}
In normalizing flows, the function $f_\phi$ must be invertible; for computational efficiency, both the inverse $f_\phi^{-1}$ and Jacobian determinant should be simple to calculate. 
In this work, we use RealNVP~\cite{dinh2016density}.

\section{Structured Normalizing Flow GP (SNFGP)}

\subsection{Structured Normalizing Flows}
By construction, normalizing flows do not reduce the dimensionality of the input data.
As in many scientific applications, the important variation across ChemCam spectra is believed to lie in a lower dimensional manifold.
Therefore, we consider structured normalizing flows that rely on first transforming the input data vector $\mathbf{y} \in \mathbb{R}^P$ to a lower dimensional representation $\mathbf{w} \in \mathbb{R}^K$ via a function $g_\theta$ before passing through a normalizing flow \cite{kontolati2021neural}.
We assume that the transformation $g_\theta$ is injective and has a pseudo-inverse $g^\dagger_\theta$ satisfying
\begin{align}
    \mathbf{y}^{(i)} \approx (g_\theta^\dagger \circ g_\theta) \left(\mathbf{y}^{(i)}\right). 
\end{align}
Under this assumption \cite{cunningham2020normalizing}, the change of variables formula for the dimension-reduced data is
\begin{align}
    p_Y(\mathbf{y}) = p_W(g_\theta(\mathbf{y})) \left| \tfrac{\partial g_\theta(\mathbf{y})}{\partial \mathbf{y}} \tfrac{\partial g_\theta(\mathbf{y})}{\partial \mathbf{y}}^T \right|^{-\tfrac{1}{2}}. \label{eq:multichangevar}
\end{align}

In this work, we let $g_\theta$ be the $K$-dimensional principal components analysis (PCA) representation. 
As long as enough principal components are retained, Eq~\ref{eq:multichangevar} will approximately hold.
If we let $g_\theta(\mathbf{y}) = \mathbf{W}^T \mathbf{y}$ where $\mathbf{W} \in \mathbb{R}^{P \times K}$ is an orthogonal basis, then $\frac{\partial g_\theta(\mathbf{y})}{\partial \mathbf{y}} = \mathbf{W}^T$ so we have 
\begin{align}
    p_Y(\mathbf{y}) = p_W(g_\theta(\mathbf{y})) \tfrac{1}{\sqrt{\left|\mathbf{W}^T \mathbf{W} \right|}}.
\end{align}
By orthogonality, $\left|\mathbf{W}^T \mathbf{W} \right| = \prod_{k=1}^K w_{kk}$ where $w_{kk}$ are the diagonal elements of $\mathbf{W}^T \mathbf{W}$.

\subsection{SNFGP Model Specification}
The SNFGP model is specified through the conditional density $p_{Y|X}(\mathbf{y}|\mathbf{x})$ which is written in terms of GP marginal likelihoods via the structured normalizing flow as follows.
We presume that the $K$-dimensional normalizing flow latent space can be modeled as a set of $K$ independent univariate-output GP models $p_{Z_k|X}(z_k | \mathbf{x})$:
\begin{align}
    p_{Z|X}(\mathbf{z} | \mathbf{x}) = p_{Z|X}(z_1, ..., z_K | \mathbf{x}) = \prod_{k=1}^K p_{Z_k|X}(z_k | \mathbf{x}).
\end{align}

Recalling that $\mathbf{w} = g_\theta(\mathbf{y}) = \mathbf{W}^T\mathbf{y}$ is the dimension reduction transform and $\mathbf{z} = f_\phi(\mathbf{w})$ the normalizing flow, we obtain the density of $\mathbf{y}$ conditional on inputs $\mathbf{x}$ by application of the change of variables formula Eq~\ref{eq:multichangevar}:
\begin{align}
    p_{Y|X}(\mathbf{y} | \mathbf{x}) = p_{Z|X}(\mathbf{z} | \mathbf{x}) \left| \frac{\partial f_\phi(\mathbf{w})}{\partial \mathbf{w}} \right|\frac{1}{\sqrt{\prod_{k=1}^K w_{kk} }} \label{eq:lik}
\end{align}
where $\mathbf{w} = g_\theta(\mathbf{y})$.
Eq~\ref{eq:lik} can then be maximized with respect to parameters $\gamma_k$ for each individual Gaussian process, in addition to the NF parameters $\phi$.

\subsection{Implementation Details}
In this work, we first estimate $g_\theta$ via PCA with $K = 15$ components (capturing over 96\% of the variance in the LIBS training set).
We then learn the NF parameters $\phi$ along with the GP parameters $\gamma$ by maximizing the transformed GP marginal likelihood in Eq~\ref{eq:lik}.
For the normalizing flow $f_\phi$, we use a RealNVP architecture with six coupling layers.
Note that $p_{Z|X}$ involves all of the data examples for evaluation; with a computational complexity of $O(N^3)$, the GP can become prohibitive for large data sets and will dominate the computational cost (since RealNVP architectures are designed for efficient forward, backward, and Jacobian evaluation). 
However, recent work suggests that mini-batch training for Gaussian process models is computationally efficient and accurate~\cite{chen2020stochastic}, so we use batch sizes of 512 data points for each update (via Adam optimizer, learning rate 0.0005).
We implement the model in Pytorch~\cite{paszke2019pytorch} with some custom layer functions from \texttt{pytorch\_flows} (\url{https://github.com/ikostrikov/pytorch-flows}).
We randomly select distinct materials for the training and test sets, but also select a set of `extrapolation regime' test materials by including all materials with SiO$_2$ composition greater than 0.9 in the test set and using only materials with composition less than 0.8 in the training set.
This results in 2,109 unique spectra corresponding to 422 unique materials in the training set, 150 spectra corresponding to 30 unique materials in the validation set, and 18 unique materials (disjoint from the training/validation sets) with a total of 90 spectra in the test set.

\subsection{Model Evaluation}
To demonstrate goodness of fit of the model, we investigate the Gaussian process predictive accuracy in $\mathbf{z}$ space on both the training and test data to determine whether the model residuals indicate lack of fit.
To assess generative model performance, we sample from the fitted Gaussian process conditional on the true test set composition values, then propagate the samples through the inverse normalizing flow and pseudo-inverse dimension reduction functions to obtain samples in the LIBS spectral space.
Across wavelengths for each spectrum, we compute the root mean squared error (RMSE), the $R^2$, and the coverage.
RMSE and $R^2$ measure how well the mean predicted spectra matches the true spectrum, while the coverage metric constructs nominal $(1-\alpha)$ uncertainty intervals using the $[\alpha/2, \, 1-\alpha/2]$ quantiles of the predicted spectral samples and estimates the proportion of wavelengths for which the true spectrum fell within the uncertainty intervals.
Finally, we ask at whether, given a new test spectrum, its latent representation $\mathbf{z}^*$ can be used to infer the corresponding composition $\mathbf{x}^*$ 
using grid search to find the maximum of the likelihood function (Eq~\ref{eq:lik}) with respect to unknown $\mathbf{x}^*$ based on the GP predictive distribution and likelihood intervals~\cite{owen1988empirical} to describe uncertainty.

\section{Results}

\subsection{Goodness of Fit}
\begin{figure}
    \centering
    \includegraphics[width=\linewidth]{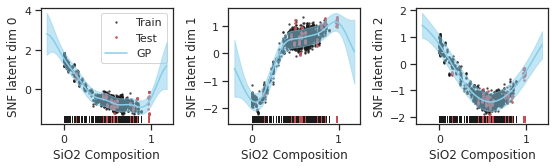}
    \caption{Gaussian process predictive distribution (mean line with shading for 95\% uncertainty interval) as a function of SiO$_2$ composition for three of the 15 normalizing flow latent dimensions. The predictive mean describes the patterns in the data accurately and the uncertainty intervals appear to capture the majority of the data.}
    \label{fig:gof}
\end{figure}

First, we investigate how well the GP model describes the data in the latent space.
Fig~\ref{fig:gof} shows, for three of the 15 total latent dimensions, scatter plots of the SiO$_2$ composition against the latent representation value.
Black points represent training data while red points represent test data; rug plots along the horizontal axis indicate the distributions of test and training data, including the `extrapolation regime' test set with SiO$_2 > 0.9$. 
The GP predictive distribution is shown as a function of composition, with the mean function prediction as a blue line and 95\% uncertainty intervals as shaded blue areas.
It appears that the GP models adequately capture variation in the latent space conditional on the composition, and the uncertainty intervals capture the data distribution well.
Some of the inherent variability in the data (reflected by the spread around the mean prediction) comes from uncontrollable sources (shot-to-shot variations for the same target) while other variability may come from the unmodeled influence of other elemental compositions.

\subsection{Generative Model Performance}
Across the test set, we evaluate how well the generative model captures the characteristics of the spectra conditional on the true composition. 
Table~\ref{tab:genmodel} gives the mean and standard deviation of the three performance metrics across all test set spectra, divided into interpolation and extrapolation regimes.
In the interpolation regime, we note that the PCA decomposition alone incurs average RMSE near 0.0001, so our model introduces some additional error on average in reconstructing the spectra, but generally the $R^2$ is high.
In the extrapolation regime, the RMSE and $R^2$ indicate worse performance (as expected, because no training data was seen in this area of input space).
For assessing coverage, we used the 2.5\% and 97.5\% quantiles to obtain uncertainty intervals (nominal 95\% coverage).
In the interpolation regime, we achieve nominal coverage, with only a slightly lower coverage in the extrapolation regime. 
This demonstrates an important property of the model: while predictions may be inaccurate when extrapolating, the uncertainty intervals expand and can therefore still contain the true data values.
\begin{table}[]
    \centering
    \begin{tabular}{l|r|r}
        & Interpolation & Extrapolation \\
        Metric & Mean (SD) & Mean (SD) \\
        \hline
        RMSE ($\times 10^{-2})$ & 0.04 (0.01) & 0.15 (0.05) \\
        $R^2$ & 0.91 (0.07) & -0.39 (0.70) \\
        Coverage & 0.95 (0.07) & 0.89 (0.04)
    \end{tabular}
    \caption{Generative model performance measured on the test set via RMSE and $R^2$ of the mean predicted spectrum and coverage of a 95\% uncertainty interval, split into interpolation and extrapolation regimes.}
    \label{tab:genmodel}
\end{table}

\begin{figure}
    \centering
    \includegraphics[width=\linewidth]{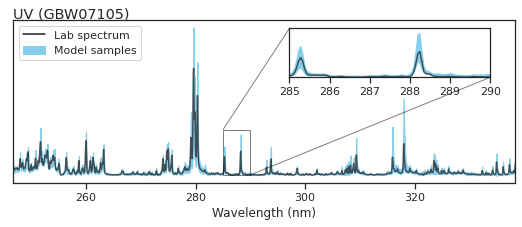}
    \caption{Generated spectra samples compared to a test set spectrum (target GBW07105) for the UV spectrometer, conditional on the true composition. Inset panel zooms in to show detail of an Si spectral line near 288.2 nm.}
    \label{fig:genspectra}
\end{figure}

Fig~\ref{fig:genspectra} shows generated spectral samples for a given input composition (SiO$_2$ oxide weight percent 44.6\%) with a test set spectrum corresponding to that composition shown in black; for simplicity, we show results only for the UV spectrometer.
The generated spectral samples appear to capture the general shape of the true spectrum. 
Zooming in on a key Si spectral line near 288.2 nm, we see some variation across model samples, but the peak appears in all samples.

\subsection{Inferring Generating Parameters}
\begin{figure}
    \centering
    \includegraphics[width=\linewidth]{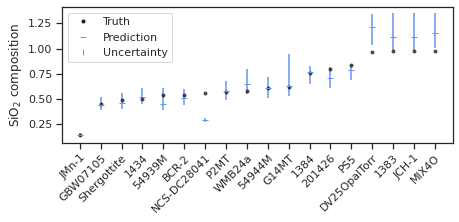}
    \caption{For the test set materials (named along the horizontal axis), we use the SNFGP likelihood to infer the composition with uncertainty. Many uncertainty intervals cover the true values. The rightmost four materials are the `extrapolation regime' test set; as expected, the predictions are less accurate, but with wider uncertainty intervals.}
    \label{fig:compmle}
\end{figure}

For a set of eighteen test set spectra representing distinct targets, we estimate the composition value corresponding to the maximum likelihood estimator (MLE) and generate asymmetric 95\% confidence intervals for the MLEs using likelihood ratio intervals.
Fig~\ref{fig:compmle} shows the MLEs (blue horizontal bars) with the uncertainty intervals (blue vertical bars) compared to the true compositions (black dots) for the different test set materials.
The intervals cover the true composition value for most of the materials, and in many cases the intervals are fairly tight.
We note that the NCS-DC28041 material appears to be an outlier in principal components space (prior to learning the SNFGP), indicating that this sample may require further investigation.
The four materials with the largest compositions correspond to the `extrapolation regime'; we note that while while the predictions are more biased, the uncertainty intervals are larger.

\section{Conclusions}
In this work, we have presented SNFGP: a novel generative model that combines dimension reduction, normalizing flows, and Gaussian process regression.
SNFGP conditions on input values to generate complex, structured, high-dimensional outputs.
We demonstrate the model on LIBS spectra from the ChemCam instrument, where the model generates realistic spectra conditional on an input composition and provides a principled way to quantify uncertainty in predictions of the input composition given a new spectral observation.
Importantly, we demonstrate that the SNFGP model has good properties when extrapolating from the training data, a property not shared by many machine learning models.
In future work, we plan to compare our method to related methods such as the GPVAE in terms of performance and computational complexity and to expand the application to include ChemCam data from Mars (including modeling the Earth/Mars data discrepancy).



\section{Acknowledgments}
This project was supported by the Laboratory Directed Research and Development program of Los Alamos National Laboratory under project number LDRD-20210043DR.

\bibliography{ref}

\end{document}